\documentclass[runningheads]{llncs}
\usepackage{graphicx}
\usepackage{comment}
\usepackage{amsmath,amssymb} 
\usepackage{color}
\usepackage{multirow}
\usepackage[export]{adjustbox}
\usepackage{extra_commands}

\begin{document}
\pagestyle{headings}
\mainmatter
\def\ECCVSubNumber{4387}  

\title{Learning Visual Commonsense \\ for Robust Scene Graph Generation}

\titlerunning{Learning Visual Commonsense for Robust Scene Graph Generation}
%
\author{Alireza Zareian\thanks{Equal contribution.} \and
Zhecan Wang\textsuperscript{$\star$} \and
Haoxuan You\textsuperscript{$\star$} \and
Shih-Fu Chang}
\authorrunning{Alireza Zareian et al.}
%
\institute{Columbia University, New York NY 10027, USA \\
\email{\{az2407,zw2627,hy2612,sc250\}@columbia.edu}}
\maketitle

\begin{abstract}

Scene graph generation models understand the scene through object and predicate recognition, but are prone to mistakes due to the challenges of perception in the wild.
Perception errors often lead to nonsensical compositions in the output scene graph, which do not follow real-world rules and patterns, and can be corrected using commonsense knowledge.
We propose the first method to acquire visual commonsense such as affordance and intuitive physics automatically from data, and use that to improve the robustness of scene understanding. To this end, we extend Transformer models to incorporate the structure of scene graphs, and train our Global-Local Attention Transformer 
on a scene graph corpus. Once trained, our model can be applied on any scene graph generation model and correct its obvious mistakes, resulting in more semantically plausible scene graphs. 
Through extensive experiments, we show our model learns commonsense better than any alternative, and improves the accuracy of state-of-the-art scene graph generation methods. 
\end{abstract}

\section{Introduction \label{sec:intro}}

\begin{figure}[t]
\centering
\includegraphics[width=1.0\linewidth]{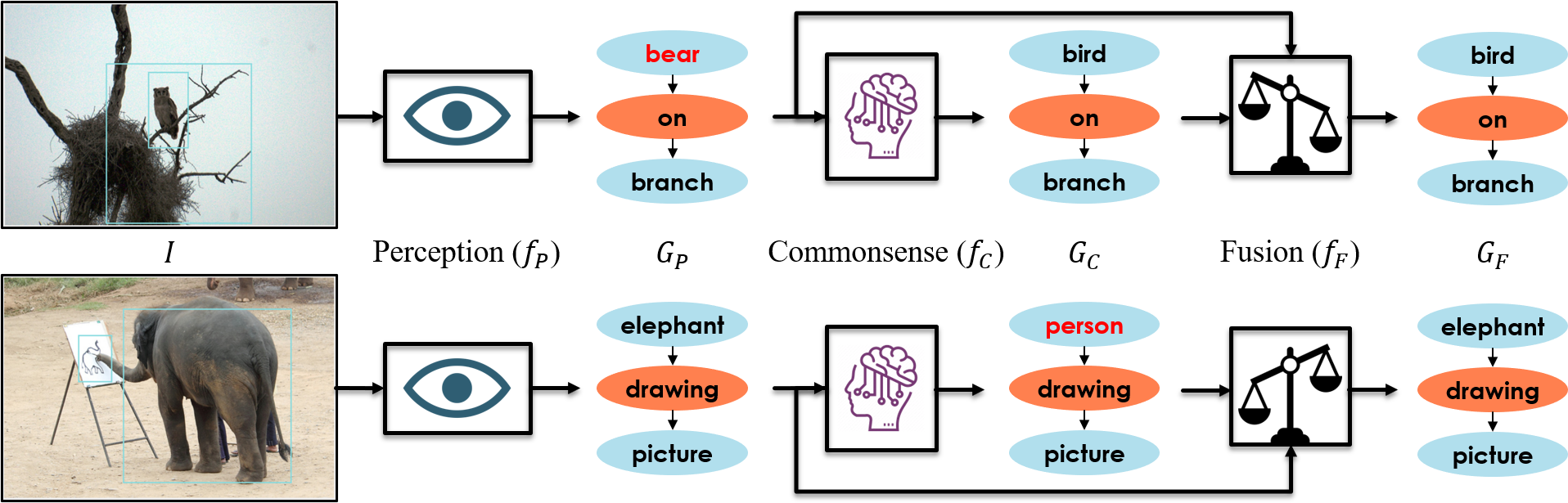}
\caption{Overview of the proposed method: We propose a commonsense model that takes a scene graph generated by a perception model and refines that to make it more plausible. Then a fusion module compares the perception and commonsense outputs and generates a final graph, incorporating both signals.} 
\label{fig:overview}
\end{figure}

In recent computer vision literature, there is a growing interest in incorporating commonsense reasoning and background knowledge into the process of visual recognition and scene understanding \cite{kato2018compositional,yu2017visual,kumar2018dock,jiang2018hybrid,zellers2019recognition}. In Scene Graph Generation (SGG), for instance, external knowledge bases \cite{gu2019scene} and dataset statistics \cite{chen2019knowledge,zellers2018neural} have been utilized to improve the accuracy of entity (object) and predicate (relation) recognition. The effect of these techniques is usually to correct obvious perception errors, and replace with more plausible alternatives. For instance, Figure \ref{fig:overview} (top) shows an SGG model mistakenly classifies a \texttt{bird} as a \texttt{bear}, possibly due to the 
dim lighting and small object size.
However, a commonsense model can correctly predict \texttt{bird}, because \texttt{bear on branch} is a less common situation, less aligned with intuitive physics, or contrary to animal behavior.

Nevertheless, existing methods to incorporate commonsense into the process of visual recognition have two major limitations. Firstly, they rely on an external source of commonsense, such as crowd-sourced or automatically mined commonsense rules, which tend to be incomplete and inaccurate~\cite{gu2019scene}, or statistics directly gathered from training data, which are limited to simple heuristics such as co-occurrence frequency~\cite{chen2019knowledge}. In this paper, we propose the first method to learn graphical commonsense automatically from a scene graph corpus, which does not require external knowledge, and \textbf{acquires} commonsense by learning complex, structured patterns beyond simple heuristics.

Secondly, most existing methods are strongly vulnerable to data bias as they integrate data-driven commonsense knowledge into data-driven neural networks. 
For instance, the commonsense model in Figure~\ref{fig:overview} mistakes the \texttt{elephant} for a \texttt{person}, in order to avoid the bizarre triplet \texttt{elephant drawing picture}, while the \texttt{elephant} is quite clear visually, and the perception model already recognizes it correctly. None of the existing efforts to equip scene understanding with commonsense have studied the fundamental question of whether to trust perception or commonsense, \textit{i.e.}, what you see versus what you expect.
In this paper, we propose a way to disentangle perception and commonsense into two separately trained models, and introduce a method to exploit the disagreement between those two models to achieve the best of both worlds.

To this end, we first propose a mathematical formalization of visual commonsense, as a problem of auto-encoding perturbed scene graphs. Based on the new formalism, we propose a novel method to learn visual commonsense from annotated scene graphs. We extend recently successful transformers \cite{vaswani2017attention} by adding local attention heads to enable them to encode the structure of a scene graph, and we train them on a corpus of annotated scene graphs to predict missing elements of a scene via a masking framework similar to BERT \cite{devlin2018bert}. As illustrated in Figure \ref{fig:glat}, our commonsense model learns to use its experience to imagine which entity or predicate could replace the mask, considering the structure and context of a given scene graph. Once trained, it can be stacked on top of any perception (\textit{i.e.,} SGG) model to correct nonsensical mistakes in the generated scene graphs. 

The output of the perception and commonsense models can be seen as two generated scene graphs with potential disagreements. We devise a fusion module that takes those two graphs, along with their classification confidence values, and predicts a final scene graph that reflects both perception and commonsense knowledge. The degree to which our fusion module trusts each input varies for each image, and is determined based on the estimated confidence of each model. This way, if the perception model is uncertain about the \texttt{bird} due to darkness, the fusion module relies on the commonsense more, and if perception is confident about the \texttt{elephant} due to its clarity, the fusion module \textit{trusts its eyes}.

We conduct extensive experiments on the Visual Genome datasets~\cite{krishna2017visual}, showing (1) The proposed GLAT model outperforms existing transformers and graph-based models in the task of commonsense acquisition; (2) Our model learns various types of commonsense that are absent in SGG models, such as object affordance and intuitive physics; (3) The proposed model is robust to dataset bias, and shows commonsensical behavior even in rare and zero-shot scenarios; (4) The proposed GLAT and Fusion mechanism can be applied on any SGG method to correct their mistakes and improve their accuracy.
The main contributions of this paper are the following:
\begin{itemize}
    \item We propose the first method for learning structured visual commonsense, Global-Local Attention Transformer (GLAT), which does not require any external knowledge, and outperforms conventional transformers and graph-based networks.
    \item We propose a cascaded fusion architecture for Scene Graph Generation, which disentangles commonsense reasoning from visual perception, and integrates them in a way that is robust to the failure of each component.
    \item We report experiments that showcase our model's unique ability of learning commonsense without picking up dataset bias, and its utility in downstream scene understanding.
\end{itemize}

\section{Related Work \label{sec:related}}

\subsection{Commonsense in computer vision}


Incorporating commonsense knowledge has been explored in various computer vision tasks such as object recognition~\cite{chen2018iterative,wang2018zero,lee2018multi}, object detection~\cite{kumar2018dock}, semantic segmentation~\cite{qi2019ke}, action recognition~\cite{kato2018compositional}, visual relation detection~\cite{yu2017visual}, scene graph generation~\cite{chen2019knowledge,zellers2018neural,gu2019scene}, and visual question answering~\cite{narasimhan2018straight,su2018learning}. There are two aspects to study about these methods: where their commonsense comes from, and how they use it.

Most methods either adopt an external curated knowledge base such as ConceptNet~\cite{speer2017conceptnet,gu2019scene,wang2018zero,lee2018multi,qi2019ke,narasimhan2018straight}, or acquire commonsense automatically by collecting statistics over an often annotated corpus~\cite{chen2018iterative,chen2019knowledge,zellers2018neural,kumar2018dock,su2018learning,yu2017visual}. Nevertheless, the former group are limited to incomplete external knowledge, and the latter are based on ad-hoc, hard-coded heuristics such as the co-occurrence frequency of categories. Our method is the first to formulate visual commonsense as a machine learning task, and train a graph-based neural network to solve it. There are a third group of works that focus on a particular type of commonsense by designing a specialized model, such as intuitive physics~\cite{groth2018shapestacks}, or object affordance~\cite{chuang2018learning}. We put forth a more general framework that includes but is not limited to physics and affordance, by exploiting scene graphs as a versatile semantic representation. The most similar to our work is~\cite{wang2020visual}, which only models object co-occurrance patterns, while we also incorporate object relationships and scene graph structure.

When it comes to utilizing commonsense, existing methods integrate it within the inference pipeline, either by retrieving a set of relevant facts from a knowledge base and feeding as additional features to the model~\cite{narasimhan2018straight,gu2019scene,su2018learning}, or by employing a graph-based message propagation process to embed the structure of the knowledge graph within the intermediate representations of the model~\cite{chen2018iterative,kato2018compositional,chen2019knowledge,wang2018zero,lee2018multi}.
Some other methods distill the knowledge during training through auxiliary objectives, making the inference simple and free of external knowledge~\cite{qi2019ke,yu2017visual}. 
Nevertheless, in all those approaches, commonsense is seamlessly infused into the model and cannot be disentangled. This makes it hard to study and evaluate commonsense and perception separately, or control their influence. Few methods have modeled commonsense as a standalone module which is late-fused into the prediction of the perception model~\cite{kumar2018dock,zellers2018neural}. Yet, we are the first to devise separate perception and commonsense models, and adaptively weigh their importance based on their confidence, before fusing their predictions.


\subsection{Commonsense in scene graph generation}

Zellers \textit{et al.} \cite{zellers2018neural} were the first to explicitly incorporate commonsense into the process of scene graph generation. They biased predicate classification logits using a pre-computed frequency prior that is a static distribution, given each entity class pair. Although this significantly improved their overall accuracy, the improvement is mainly due to the fact that they favor frequent triplets over others, which is statistically rewarding. Even if their model classifies the relation between a \texttt{person} and a \texttt{hat} as \texttt{holding}, their frequency bias would most likely change that to \texttt{wearing}, which is more frequent.

More recently, Chen \textit{et al.} \cite{chen2019knowledge} employed a less explicit way to incorporate the frequency prior within the process of entity and predicate classification. They embed the frequencies into the edge weights of their inference graph, and utilize those weights within their message propagation process. This improves the results especially on less frequent predicates, since it less strictly enforces the statistics on the final decision. However, this way commonsense is integrated implicitly into the SGG model and cannot be probed or studied in isolation. We remove the adverse effect of statistical bias while keeping the commonsense model disentangled from perception.

Gu \textit{et al.}~\cite{gu2019scene} exploits ConceptNet~\cite{speer2017conceptnet} rather than dataset statistics, which is a large-scale knowledge graph comprising relational facts about concepts, \textit{e.g.} \texttt{dog is-a animal} or \texttt{fork is-used-for eating}. Given each detected object, they retrieve ConceptNet facts involving that object class, and employ a recurrent neural net and an attention mechanism to encode those facts into the object features, before classifying objects and predicates. Nevertheless, ConceptNet is not exhaustive, since it is extremely hard to
compile all commonsense facts. Our method does not depend on a limited source of external knowledge, and acquires commonsense automatically, via a generalizable neural network. 


\subsection{Transformers and graph-based neural networks}

Transformers were originally proposed to replace recurrent neural networks for machine translation, by stacking several layers of multi-head attention~\cite{vaswani2017attention}. Ever since, transformers have been successful in various vision and language tasks~\cite{devlin2018bert,wang2018non,lu2019vilbert}. 
Particularly, BERT~\cite{devlin2018bert} randomly replaces some words from a given sentence with a special \texttt{MASK} token and tries to reconstruct those words. Through this self-supervised game, BERT acquires natural language, and can transfer its language knowledge to perform well in other NLP tasks. 
We use a similar self-supervised strategy to learn to complete missing pieces of a scene graph. Rather than language, our model acquires the ability to imagine a scene in a structured, semantic way, which is a hallmark of human commonsense.

Transformers treat their input as a set of tokens, and discard any form of structure among them. To preserve the order of tokens in a sentence, BERT augments the initial embedding of each token with a position embedding before feeding into transformers.
Scene graphs, on the other hand, have a more complex structure that cannot be embedded in such a trivial way. Recently, Graph-based Neural Networks (GNN) have been successful to encode graph structures into node representations, by applying several layers of neighborhood aggregation. 
More specifically, each layer of a GNN represents each node by a trainable function that takes the node as well as its neighbors as input. Graph convolutional nets~\cite{kipf2016semi}, gated graph neural nets~\cite{li2015gated}, and graph attention nets~\cite{velivckovic2017graph} all implement this idea with different computational models for neighborhood aggregation. GNNs have been widely utilized for scene graph generation by incorporating context~\cite{xu2017scene,yang2018graph,zareian2020weakly}, but we are the first to exploit GNNs to learn visual commonsense.

We adopt graph attention nets due to their similarity to transformers in using attention. 
The main difference of graph attention nets to transformers is that instead of representing each node by an attention over all other nodes, they only compute an attention over immediate neighbors. Inspired by that, we use a BERT-like transformer network, but replace half of its attention heads by local attention, simply by enforcing the attention between non-neighbor nodes to zero. Through ablation experiments in Section~\ref{sec:experiments}, we show the proposed Global-Local Attention Transformers (GLAT) outperforms conventional transformers, as well as widely used graph-based models such as graph convolution nets and graph attention nets.

\section{Method \label{sec:method}}

In this section, we first formalize the task, and propose a novel formulation of visual commonsense in connection with visual perception. We then provide an overview of the proposed architecture (Figure~\ref{fig:overview}), followed by an in-depth description of each proposed module.

We define a scene graph as $G=(\mathcal{N}_e, \mathcal{N}_p, \mathcal{E}_s, \mathcal{E}_o)$, where $\mathcal{N}_e$ is a set of entity nodes, $\mathcal{N}_p$ is a set of predicate nodes, $\mathcal{E}_s$ is a set of edges from each predicate to its subject (which is an entity node), and $\mathcal{E}_o$ is a set of edges from each predicate to its object (that also is an entity node). Each entity node is represented with an entity class $c_e \in \mathcal{C}_e$ and a bounding box $b \in [0,1]^4$, while each predicate node is represented with a predicate class $c_p \in \mathcal{C}_p$ and is connected to exactly one subject and one object. Note that this formulation of scene graph is slightly different from the conventional one \cite{xu2017scene}, as we formulate predicates as nodes rather than edges. This tweak does not cause any limitation since every scene graph can be converted from the conventional representation to our representation. However, this formulation allows multiple predicates between the same pair of entities, and it also enables us to define a unified attention over all nodes no matter entity or predicate. 

Given a training dataset with many images $I\in [0,1]^{h\times w\times c}$ paired with ground truth scene graphs $G_T$, our goal is to train a model that takes a new image and predicts a scene graph that maximizes $p(G|I)$. This is equivalent of maximizing $p(I|G)p(G)$, which breaks the problem into what we call \textit{perception} and \textit{commonsense}. 
In our proposed intuition, commonsense is the mankind's ability to predict which situations are possible and which are not, or in other words, what makes \textit{sense} and what does not. This can be seen as a prior distribution $p(G)$ over all possible situations in the world, represented as scene graphs.  
Perception, on the other hand, is the ability to form symbolic belief from raw sensory data, which are respectively $G$ and $I$ in our case. Although the goal of computer vision is to solve the Maximum a Posteriori (MAP) problem (maximizing $p(G|I)$), neural nets often fail to estimate the posterior, unless the prior is explicitly enforced in the model definition~\cite{malinin2018predictive}. This is while in computer vision, the prior is often overlooked, or inaccurately considered to be a uniform distribution, making MAP equivalent to Maximum Likelihood (ML), \textit{i.e.}, finding $G$ that maximizes $p(I|G)$~\cite{romaszko2017vision}. 

We propose the first method to explicitly approximate the MAP inference by devising an explicit prior model (commonsense). Since posterior inference is intractable, we propose a two-stage framework as an approximation:
We first adopt any off-the-shelve SGG model as the \textit{perception model}, which takes an input image and produces a perception-driven scene graph, $G_P$, that approximately maximizes the likelihood.
Then we propose a \textit{commonsense model}, which takes $G_P$ as input, and produces a commonsense-driven scene graph, $G_C$, to approximately maximize the posterior, i.e.,
\begin{align}
G_P = f_P(I) &\approx \argmax_G p(I|G),\\
G_C = f_C(G_P) &\approx \argmax_G p(I|G)p(G)
,\end{align}
where $f_P$ and $f_C$ are the perception and commonsense models. 
The commonsense model can be seen as a graph-based extension of denoising autoencoders~\cite{vincent2008extracting}, which evidently can learn the generative distribution of data~\cite{kingma2013auto,alain2014regularized}, that is $p(G)$ in our case. Accordingly, $f_C$ can take any scene graph as input and produce a more plausible graph by only slightly changing the input. 
A key design choice here is the fact that $f_C$ does not take the image as input. Otherwise, it would be hard to ensure it is purely learning commonsense and not perception.

Ideally, $G_C$ is the best decision to make, since it maximizes the posterior distribution. However, in practice autoencoders tend to under-represent long-tailed distributions and only capture the modes.
This means the commonsense model may fail to predict less common structures, in favor of more statistically rewarding alternatives. To alleviate this problem, we propose a \textit{fusion module} that takes $G_P$ and $G_C$ as input, and outputs a fused scene graph, $G_F$, which is the final output of our system. This can be seen as a decision-making agent that has to decide how much to trust each model, based on how confident they are. 

Figure~\ref{fig:overview} illustrates an overview of the proposed architecture. In the rest of this section, we elaborate each module in detail.

\subsection{Global-Local Attention Transformers}

\begin{figure}[t]
\centering
\includegraphics[width=1.0\linewidth]{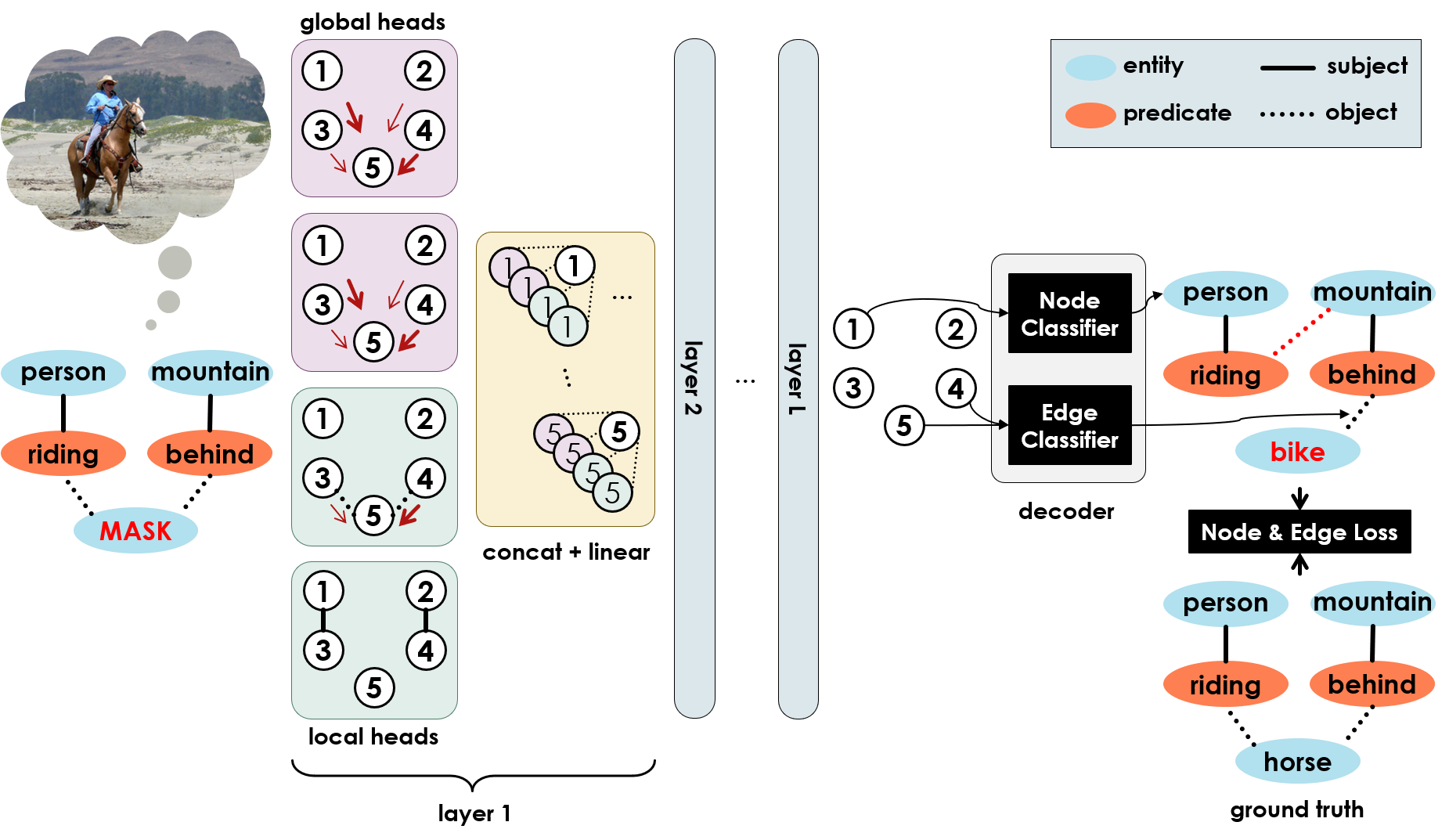}
\caption{The proposed Global-Local Attention Transformer (GLAT), and its training framework: We augment transformers with local attention heads to help them encode the structure of scene graphs within node embeddings. The decoder takes the embeddings of a perturbed scene graph and reconstructs the correct scene graph without having access to the image. Note this figure only shows the commonsense block of our overall pipeline shown in Figure~\ref{fig:overview}.}
\label{fig:glat}
\end{figure}

We propose the first graph-based visual commonsense model, which learns a generative distribution over the semantic structure of real-world scenes, through a denoising autoencoder framework.
Inspired by BERT~\cite{devlin2018bert}, which reconstructs masked tokens in a sentence through stacked layers of multi-head attention, we propose Global-Local Attention Transformers (GLAT) that take a graph with masked nodes as input, and reconstructs the missing nodes. 
Figure~\ref{fig:glat} illustrates how GLAT works.
Given an input scene graph $G_P$, we represent node $i$ as a one-hot vector $x_i^{(0)}$, that includes entity and predicate categories, as well as a special \texttt{MASK} class. We stack node representations as rows of a matrix $X^{(0)}$ for notation purposes. 

GLAT takes $X^{(0)}$ as input and represents each node by encoding the structure and context. To this end, it applies $L$ layers of multi-head attention on the input nodes. Each layer $l$ creates new node representations $X^{(l)}$, by applying a linear layer on the concatenated output of that layer's attention heads. More specifically,
\begin{equation}
\begin{aligned}
X^{(l)} = \concat_{h \in \mathcal{H}_l} \Big[ h(X^{(l-1)}) \Big] \times W_l + b_l
,\end{aligned}
\end{equation}
where $\mathcal{H}_l$ is the set of attention heads for layer $l$, $W_l$ and $b_l$ are trainable fusion weights and bias for that layer, and the concatenation operates along columns.
We use two types of attention head, namely global and local. Each node can attend to all other nodes through global attention, while only its neighbors through local attention. We further divide local heads based on the type of edge they use, in order to differentiate the way subjects and objects interact with predicates, and vice versa. Therefore, we can write:
\begin{equation}
\begin{aligned}
\mathcal{H}_l = \mathcal{H}_l^G \cup \mathcal{H}_l^{LS} \cup \mathcal{H}_l^{LO}
.\end{aligned}
\end{equation}
All heads within each subset are identical, except they have distinct parameters that are initialized and trained independently. Each global head $h^G$ operates as a typical self-attention would:
\begin{equation}
\begin{aligned}
h^G(X) = \big[ q(X)^T k(X) \big] v(X)
,\end{aligned}
\end{equation}
where $q,k,v$ are query, key, and value heads, each a fully connected network, typically (but not necessarily) with a single linear layer.
A local attention is the same, except queries can only interact with keys of their immediate neighbor nodes. For instance in subject heads,
\begin{equation}
\begin{aligned}
h^{LS}(X) = \big[ q(X)^T k(X) \odot A_s \big] v(X)
,\end{aligned}
\end{equation}
where $A_s$ is the adjacency matrix of subject edges, which is 1 between from each predicate to its subject and vice versa, and 0 elsewhere. We similarly define $A_o$ and $h^{LO}$ for object edges.

Once we get contextualized, structure-aware representations $x_i^{(L)}$ for each node $i$, we devise a simple decoder to generate the output scene graph $G_C$, using a fully connected network that classifies each node to an entity or predicate class, and another fully connected network that classifies each pair of nodes into an edge type (subject, object or no edge). We train the encoder and decoder end-to-end, by randomly adding noise to annotated scene graphs from Visual Genome, feeding the noisy graph to GLAT, reconstructing nodes and edges, and comparing each with the original scene graph before perturbation. We train the network using two cross-entropy loss terms on the node and edge classifiers. The details of training including the perturbation process are explained in Section~\ref{sec:details}.

\subsection{Fusing Perception and Commonsense}

The perception and commonsense models each predict the output node categories using a classifier that computes a probability distribution over all classes by applying a softmax on its logits. The class with highest probability is chosen and assigned a confidence score equal to its softmax probability. More specifically, node $i$ from $G_P$ has a logit vector $L^P_i$ that has $|\mathcal{C}_e|$ or $|\mathcal{C}_p|$ dimensions depending of whether it is an entity node or predicate node. Similarly node $i$ from $G_C$ has a logit vector $L^C_i$. Note that these two nodes correspond to the same entity or predicate in the image, since GLAT does not change the order of nodes. Then the confidence of each node can be written as 
\begin{equation}
\begin{aligned}
q^P_i = \max_j \frac{\exp(L^P_i[j])}{\sum_k \exp(L^P_i[k])}
,\end{aligned}
\end{equation}
and similarly $q^C_i$ is defined given $L^C_i$. 

The fusion module takes each node of $G_P$ and the corresponding node of $G_C$, and computes a new logit vector for that node, as a weighted average of $L^P_i$ and $L^C_i$. The weights determine the contribution of each model in the final prediction, and thus have to be proportional to the confidence of each model. Therefore, we compute the fused logits as:
\begin{equation}
\begin{aligned}
L^F_i = \frac{q^P_i L^P_i + q^C_i L^C_i}{q^P_i + q^C_i}
.\end{aligned}
\end{equation}
Finally, a softmax is applied on $L^F_i$ to compute the final classification distribution for node $i$.

\section{Experiments \label{sec:experiments}}

In this section, we describe our experiments on the Visual Genome (VG) dataset in detail. We first evaluate how well our GLAT model learns visual commonsense, by comparing it to other models on the task of masked scene graph reconstruction. Then we provide a statistical analysis of our model prediction to show the kinds of commonsense knowledge it acquires, and distinguish it from bias. Next, we evaluate how effective GLAT and our fusion mechanism are for the downstream task of SGG, when applied on various perception models. We also provide several examples of how the commonsense model corrects the perceived output, and how the fusion model combines the two. 

\subsection{Implementation details\label{sec:details}}

We train the perception and commonsense models separately using the ground truth scene graphs $G_T$ from VG \cite{krishna2017visual}, particularly the version most widely used for SGG~\cite{xu2017scene}, which has 150 entity and 50 predicate classes. We then stack commonsense on top of perception and fine-tune it on VG, this time with actual scene graphs generated by perception, to adapt to the downstream task. The fusion module does not have trainable parameters and is thus only used during inference. 
We use the 75k VG scene graphs for training all models, and use the other 25k for test. We hold a small portion of the train set for validation.
Our GLAT model (and other baselines when applicable) have 6 layers, each with 8 attention heads, and has a 300-dimensional representation for each node. 
While training GLAT, we randomly mask 30\% of the nodes, which is the average number of nodes mistaken by a typical SGG model. We average the classification loss over all nodes and edges classified by the decoder, no matter masked or not. 
For fine-tuning and inference, we prune the output of the perception model before feeding to GLAT, by keeping the top 100 most confident predicates and all entities connected to those.

\subsection{Evaluating commonsense\label{sec:ablation}}

Once GLAT is trained, we evaluate it on the same task of reconstructing ground truth VG graphs that are perturbed by randomly masking 30\% of their nodes. We evaluate the accuracy of our model in classifying the masked nodes, and report the accuracy (Table~\ref{table:ablation}) separately for entity nodes and predicate nodes, as well as overall. This is a good measure of how well the model has learned commonsense, because it mimics mankind's ability to imagine what would a real-world scene look like, given some context. In Figure~\ref{fig:glat}, for instance, given the fact that there is a person riding something that is masked, we can immediately tell it is probably a bike, a motorcycle, or a horse. If we also know there is a mountain behind the masked object, and the masked object has a face and legs (not shown in the figure for brevity), then we can more certainly imagine it is a horse. By incorporating the global context of the scene, as well as the local structure of the graph, GLAT is able to effectively imagine the scene and predict the class of the entity or predicate that was masked, at a significantly higher accuracy compared to all baselines.

More specifically, we compare GLAT to: (1) A transformer~\cite{devlin2018bert} that is the same as our model, except it only has global heads; (2) A Graph Attention Net~\cite{velivckovic2017graph} which is also the same as our model, but only with local heads; and (3) A Graph Convolutional Network~\cite{kipf2016semi}, which has only one local head at each layer, and the attention is fixed to be equal for all neighbors of each node. We also compare our method with the frequency prior used by Zellers \textit{et al.}~\cite{zellers2018neural}, which can only be applied for masked predicates, and simply predicts the most frequent predicate given its subject and object. As Table~\ref{table:ablation} shows, our method significantly outperforms all aforementioned baselines, which are a good representative of any existing method to learn semantic graph reconstruction. 

To provide a better sense of the commonsense knowledge our model learns, we apply GLAT on the entire VG test set, using the procedure detailed below (Section~\ref{sec:sggeval}), and collect its prediction statistics in a diverse set of situations. We elaborate using an example, shown in the top left cell of Table~\ref{table:comse_examples}. Out of all triplets from all scene graphs produced by our model, we collect those triplets that match the certain template of \texttt{person [X] horse}, and show our sorted top 5 predictions in terms of frequency. The 5 predicates most often predicted by our method between a \texttt{person} and a \texttt{horse} are \texttt{on}, \texttt{riding}, \texttt{near}, \texttt{watching}, and \texttt{behind}. These are all possible interactions between a person and a horse, and all follow the affordance properties of both \texttt{person} and \texttt{horse}. Nevertheless, when we get the same statistics from the output of a state-of-the-art scene graph generation model (IMP~\cite{xu2017scene}), we observe that it frequently predicts \texttt{person wearing horse}, which does not follow the affordance of \texttt{horse}. This can be attributed to the high frequency of \texttt{wearing} in VG annotation, which biases the IMP model, while our commonsense model is prone to such bias, and has learned affordances through the self-supervised training framework.

\begin{table}[t]
\begin{center}
\caption{\label{table:ablation} Ablation study on Visual Genome. All numbers are in percentage, and graph constraint is enforced}
\begin{tabular}{l|c c c}
\hline
Method & Entity & Predicate & Both \\
\hline\hline
Triplet Frequency~\cite{zellers2018neural} & - & 44.4 & - \\
Graph Convolutional Nets~\cite{kipf2016semi} (local-only, fixed attention) & 8.7 & 43.4 & 19.7 \\
Graph Attention Nets~\cite{velivckovic2017graph} (local-only) & 12.0 & 45.0 & 22.3 \\
Transformers~\cite{devlin2018bert} (global-only) & 14.0 & 42.3 & 22.9 \\
\textbf{Global-Local Attention Transformers (ours)} & \textbf{22.3} & \textbf{60.7} & \textbf{34.4} \\
\hline
\end{tabular}
\end{center}
\end{table}

Table~\ref{table:comse_examples} provides several more scenarios like this, demonstrating our proficiency in three types of commonsense: object affordance, intuitive physics, and object composition. As an example of physics, we choose the triplet template \texttt{[X] under bed}, and show that our model predicts plausible objects such as \texttt{pot}, \texttt{shoe}, \texttt{drawer}, \texttt{book}, and \texttt{sneaker}. This is while IMP predicts \texttt{bed under bed}, \texttt{counter under bed}, and \texttt{sink under bed}, which are all physically counter-intuitive. More interestingly, one of our frequent predictions, \texttt{book under bed}, is a composition that does not exist in training data, suggesting the knowledge acquired by GLAT is not merely a biased memory of frequent compositions in training data. 

The last type of commonsense in our illustration is object composition, \textit{i.e.}, the fact that certain objects are physical parts of other objects. For \texttt{[X] has ear}, we predict \texttt{head}, \texttt{cat}, \texttt{elephant}, \texttt{zebra}, and \texttt{person}, out of which \texttt{head has ear} and \texttt{person has ear} are not within the 10 most frequent triplets in training data that match the template. Yet our model frequently predicts them, demonstrating its unbiased knowledge. Not to mention, 4 out of 5 top predictions made by IMP are nonsensical.

\begin{table}[t]
\begin{center}
\caption{\label{table:comse_examples} Prediction statistics of our method compared to IMP~\cite{xu2017scene} in various situations, showcasing our model's commonsense knowledge, and its robustness to dataset bias. Each row is designated for a certain type of commonsense, and has three examples in three pairs of columns. Each pair of columns show the top 5 most frequent triplets matching a certain template from our model's prediction, compared to IMP. \textbf{Black} triplets are commonsensically correct, \textbf{red} triplets are wrong, \textbf{blue} are commonsensically correct but statistically rare in training data, and \textbf{green} are correct but never seen in training data.}
\includegraphics[width=1.0\linewidth]{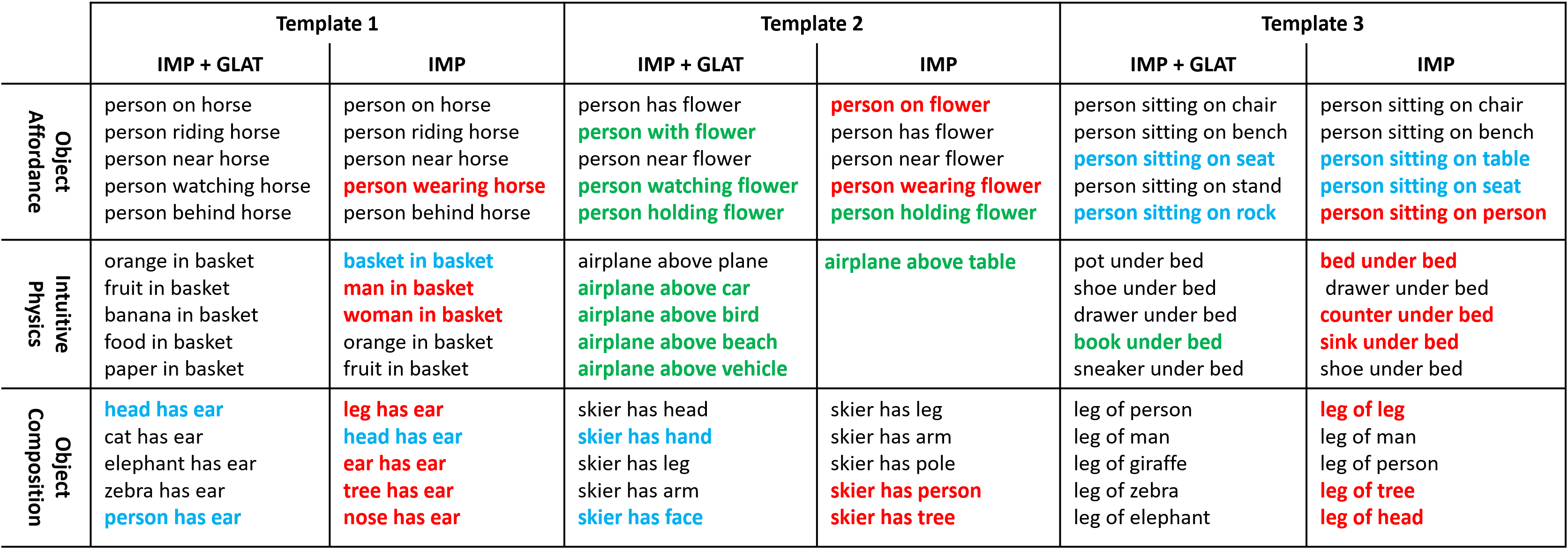}
\end{center}
\end{table}

\subsection{Evaluating scene graph generation\label{sec:sggeval}}

Now that we showed the efficacy of GLAT in learning visual commonsense and correcting perturbed scene graphs, we apply and evaluate it on the downstream task of scene graph generation. We adopt existing SGG models as our perception model, and compare their output $G_P$, to the ones corrected by our commonsense model $G_C$, as well as the final output of our system after fusion $G_F$. We compare those 3 outputs for 3 different choices of perception model, all of which have competitive state-of-the-art performance. More specifically, we use Iterative Message Passing (IMP~\cite{xu2017scene}) as a strong baseline that is not augmented by commonsense. We also use Stacked Neural Motifs (SNM~\cite{zellers2018neural}) that late-fuse a frequency prior with their output, and Knowledge-Embedded Routing Networks (KERN~\cite{chen2019knowledge}) that encode frequency prior within their internal message passing. 


To evaluate, we conventionally compute the mean recall of the top 50 (mR@50) and top 100 (mR@100) triplets predicted by each model. Each triplet is considered correct if the subject, predicate, and object are all classified correctly, and the bounding box of the subject and object have more than 50\% overlap (intersection over union) with the ground truth. We compute the recall for the triplets of each predicate class separately, and average over classes. 
The aforementioned metrics are measured in 2 sub-tasks: (1) \textsc{SGCls} is the main scenario where we classify entities and predicates given annotated bounding boxes. This way the performance is not limited by proposal quality. (2) \textsc{PredCls} provides the model with ground truth object labels, which helps evaluation focus on predicate recognition accuracy. 
Table~\ref{table:mR} shows the full comparison of all methods on all metrics. We observe that GLAT improves the performance of IMP which does not have commonsense, but does not significantly change the performance of SNM and KERN which already use dataset statistics. However, our full model which uses both the output of the perception model as well as commonsense model consistently improves SGG performance. 
In the supplementary material, we provide a more detailed analysis by breaking the results down into subgroups based on triplet frequency, and showing our performance boost is consistent in frequent and rare situations.



\begin{table}[t]
\begin{center}
\caption{\label{table:mR} The mean recall of our method compared to the state of the art on the task of scene graph generation, evaluated on the Visual Genome dataset~\cite{xu2017scene}, following the experiment settings of \cite{zellers2018neural}. All baseline numbers were borrowed from \cite{chen2019knowledge}, and all numbers are in percentage}
\begin{tabular}{l|c c | c c}
\hline
\multirow{2}{*}{Method} & \multicolumn{2}{c|}{\textsc{PredCls}} & \multicolumn{2}{c}{\textsc{SGCls}}  \\
& mR@50 & mR@100 & mR@50 & mR@100  \\
\hline\hline
IMP~\cite{xu2017scene} & 9.8 & 10.5 & 5.8 & 6.0  \\
IMP + GLAT & 11.1 & 11.9 & 6.2 & 6.5 \\
IMP + GLAT + Fusion & \textbf{12.1} & \textbf{12.9} & \textbf{6.6} & \textbf{7.0} \\
\hline
SNM~\cite{zellers2018neural} & 13.3 & 14.4 & 7.1 & 7.5  \\
SNM + GLAT & 13.6 & 14.6 & 7.3 & 7.8  \\
SNM + GLAT + Fusion & \textbf{14.1} & \textbf{15.3} & \textbf{7.5} & \textbf{7.9}  \\
\hline
KERN~\cite{chen2019knowledge} & 17.7 & 19.2 & 9.4 & 10.0  \\
KERN + GLAT & 17.6 & 19.1 & 9.3 & 10.0  \\
KERN + GLAT + Fusion & \textbf{17.8} & \textbf{19.3} & \textbf{9.9} & \textbf{10.4}  \\
\hline
\end{tabular}
\end{center}
\end{table}

\begin{figure}[t]
\centering
\includegraphics[width=1.0\linewidth]{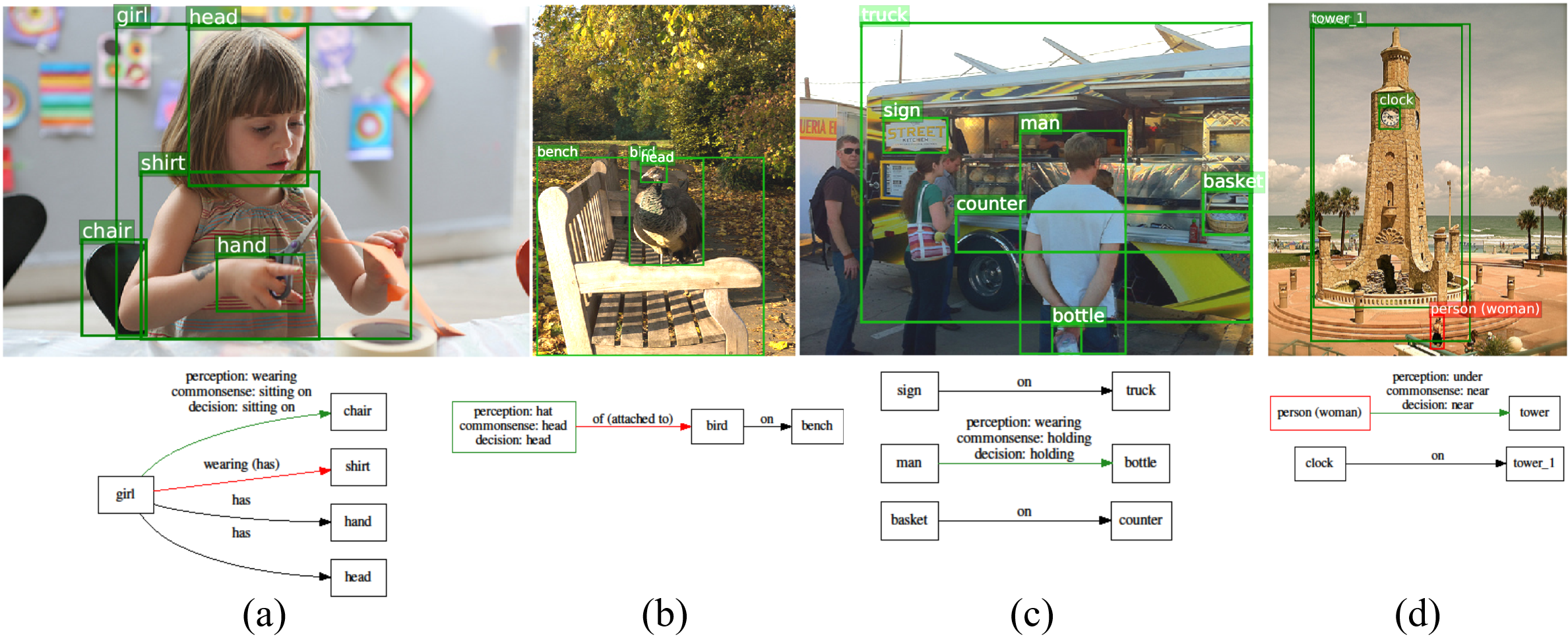}
\caption{Example scene graphs generated by the perception, commonsense, and fusion modules, merged into one graph. Entities are shown as rectangular nodes and predicates are shown as directed edges from subject to object. For entities and predicates that are identically classified by the perception and commonsense model, we simply show the predicted label. But in cases where the perception and commonsense models disagree, we show both of their predictions as well as the final output chosen by the fusion module. We show mistakes in red, with the ground truth in parentheses.}
\label{fig:example}
\end{figure}

Finally, we provide several examples in Figure~\ref{fig:example} to illustrate how our commonsense model fixes perception errors in difficult scenarios, and improves the robustness of our model. To save space, we merge the three scene graphs predicted by the perception, commonsense, and fusion models into a single graph, and emphasize any node or edge where these three models disagree. 
In example (a), the \texttt{chair} is not fully visible, and the visible part does not visually show the action of \texttt{sitting}, thus the perception model incorrectly predicts \texttt{wearing}, which is likely to be also affected by the bias due to the prevalence of \texttt{wearing} annotations in Visual Genome. However, it is trivial for the commonsense model that the affordance of \texttt{chair} is \texttt{sitting}. The fusion module correctly prefers the output of the commonsense model, due to its higher confidence. 
In (b), the perception model mistakes the \texttt{head} of the \texttt{bird} for a \texttt{hat}, due to the complexity of the lighting and the similarity of foreground and background colors. This might be also affected by the bias of \texttt{head} instances in VG, which are usually human heads, and the fact that \texttt{hat} instances typically co-occur with a \texttt{head}. Nevertheless, our commonsense model has the knowledge of object composition and knows \texttt{brid}s typically have \texttt{head}s but
not \texttt{hat}s.
Example (c) is an unusual case of \texttt{holding}, in terms of visual attributes such as arm pose. Hence, the perception model fails to predict \texttt{holding} correctly, while our commonsense model corrects that mistake by incorporating the affordance of \texttt{bottle}.
Finally, in (d), the \texttt{person} is perceived \texttt{under} the \texttt{tower} due to the camera angle, but for the commonsense model that is unlikely due to intuitive physics. Hence, it corrects the mistake and the fusion module accepts that fix.
More examples are provided in the supplementary material.

\section{Conclusion \label{sec:conclusion}}

We presented the first method to learn visual commonsense automatically from a scene graph corpus. Our method learns structured commonsense patterns, rather than simple co-occurrence statistics, through a novel self-supervised training strategy. Our unique way of augmenting transformers with local attention heads significantly outperforms transformers, as well as widely used graph-based models such as graph convolutional nets. Furthermore, we proposed a novel architecture for scene graph generation, which consists of two individual models, perception and commonsense, which are trained differently, and can complement each other under uncertainty, improving the overall robustness. To this end, we proposed a fusion mechanism to combine the output of those two models based on their confidences, and showed our model correctly determines when to trust its perception and when to fall back on its commonsense. Experiments show the effectiveness of our method for scene graph generation, and encourage future work to apply the same methodology on other computer vision tasks.

\noindent\textbf{Acknowledgement}
This work was supported in part by Contract N6600119C4032 (NIWC and DARPA). The views expressed are those of the authors and do not reflect the official policy of the Department of Defense or the U.S. Government.

\clearpage

\bibliographystyle{splncs04}
\bibliography{egbib}

\clearpage

\appendix

\title{Supplementary Material}
\titlerunning{Learning Visual Commonsense for Robust Scene Graph Generation}
\author{}
\authorrunning{Alireza Zareian et al.}
\institute{}
\maketitle

In the following, we provide additional details and analysis that was not included in the main paper due to limited space. We first provide more implementation details, followed by a novel evaluation protocol that reveals insightful statistics about the data and the state of the art performance. We finally provide more qualitative examples to showcase what our commonsense model learns.

\section{Implementation Details \label{sec:imp_details}}

We have three training stages: perception, commonsense, and joint fine-tuning. We train the perception model by closely following the implementation details of each method we adopt \cite{xu2017scene,zellers2018neural,chen2019knowledge}, with the exception that for IMP~\cite{xu2017scene}, we use the implementation by Zellers \textit{et al.}~\cite{zellers2018neural}, because it performs much better. We separately train the commonsense model (GLAT) once, independent of the perception model, as we described in the main paper. Then we stack GLAT on top of each perception model and perform fine-tuning, without the fusion module. Finally, we add the fusion model for inference.

Our GLAT implementation has 6 layers each with 8 attention heads, each with 300-D representations. We train it with a 30\% masking rate on Visual Genome (VG) training scene graphs, using an Adam optimizer with a learning rate of 0.0001, for 100 epochs. To stack the trained GLAT model on top of a perception model, we take the scene graph output of the perception model for a given image, keep the top 100 most confident triplets and remove the rest, and represent each remaining entity and predicate with a one-hot vector that specifies the top-1 predicted class. We intentionally discard the class distribution predicted by the perception model, to let the commonsense model reason independently in an abstract, symbolic space. Perception confidence is later taken into account by our fusion module. 

The resulting one-hot graph is represented in the same way as a VG graph, that we have pretrained GLAT on, and is fed into GLAT without masking any node. The GLAT decoder predicts new classes for each node, and new edges, but we ignore the new edges and keep the structure fixed. Hence, the output of GLAT looks like the output of the perception model with the exception that the classification logits of each node are changed. We perform 25 epochs of fine-tuning with 0.00001 learning rate and Adam, using the same entity and predicate loss that is typically used to train SGG models~\cite{chen2019knowledge}. 

\section{Quantitative Evaluation \label{sec:eval_details}}

\begin{table}[t]
\begin{center}
\caption{\label{table:neweval} Performance comparison in various levels of triplet frequency, in terms of R@100 (\%) for \textsc{SGCls}. \# and \% stand for absolute and relative frequency respectively.}
\begin{tabular}{l|c c c c c c|c}
\hline
Statistic/Method & \multicolumn{6}{c|}{Frequency Bins} & Average \\
\hline
\# instances in train data & 1-3 & 4-9 & 10-27 & 28-81 & 82-243 & 243-369 & - \\
\# unique triplets in bin & 86247 & 21994 & 4937 & 766 & 89 & 4 & -\\
\% unique triplets in bin  & 75.6 & 19.3 & 4.3 & 0.7 & 0.1 & 0.00004 & - \\
Total \% of test data & 14.7 & 21.0 & 28.0 & 22.5 & 10.6 & 3.1 & - \\
\hline\hline
IMP~\cite{xu2017scene} & 13.2 & 23.0 & 34.8 & 45.7 & 58.2 & 78.2 & 36.1  \\
IMP + GLAT & 13.2 & 23.0 & 34.8 & 45.9 & 58.5 & 78.7 & 37.0 \\
IMP + GLAT + Fusion & \textbf{13.3} & \textbf{23.1} & \textbf{35.0} & \textbf{46.2} & \textbf{58.7} & \textbf{79.2} & \textbf{37.4} \\
\hline
SNM~\cite{zellers2018neural} & 15.0 & 24.8 & 36.6 & 48.4 & 58.4 & 74.9 & 37.8 \\
SNM + GLAT & 15.1 & 24.8 & 36.6 & 49.0 & 58.4 & 75.2 & 37.9 \\
SNM + GLAT + Fusion & \textbf{15.1} & \textbf{24.8} & \textbf{36.7} & \textbf{49.5} & \textbf{58.5} & \textbf{75.3} & \textbf{38.0} \\
\hline
KERN~\cite{chen2019knowledge} & 16.7 & 26.2 & 37.5 & 48.4 & 59.6 & 77.1 & 38.8 \\
KERN + GLAT & 16.7 & 26.6 & 37.5 & 48.4 & 59.6 & 77.6 & 38.8 \\
KERN + GLAT + Fusion & \textbf{16.7} & \textbf{26.8} & \textbf{37.5} & \textbf{48.5} & \textbf{59.7} & \textbf{78.1} & \textbf{38.8} \\
\hline
\end{tabular}
\end{center}
\end{table}

Here we revisit the conventional evaluation process in the SGG literature, analyze its limitations, and provide an alternative to use in future work. Xu \textit{et al.}~\cite{xu2017scene} originally used overall recall (R@50 and R@100), which means for each image, they get the top 50 (or 100) triplets predicted by their model, compare to the ground truth triplets of that image, compute recall (number of matched triplets divided by the number of ground truth triplets), and average over all images. Later Chen \textit{et al.}~\cite{chen2019knowledge} revealed that since ground truth triplets in Visual Genome (both in train and test splits) have highly disproportional statistics, overall recall does not necessarily measure the usefulness of the model. In fact, a simple heuristic that always predicts the most frequent relationship for each pair of object classes, based on a fixed lookup table computed over training data (frequency baseline in \cite{zellers2018neural}), performs not much worse than the state of the art of that time, MotifNet~\cite{zellers2018neural}. 

Chen \textit{et al.}~\cite{chen2019knowledge} proposed an alternative metric, mean recall (mR@50 and mR@100), in which ground truth triplets are divided into 50 bins, based on their predicate type, the recall is computed for each bin separately, averaged over images, and then averaged over the 50 bins. This way, frequent predicates do not dominate the performance, and simplistic models are not praised for merely picking up bias.

Nevertheless, mean recall does not completely solve the imbalance problem, since even within each bin (predicate type), some compositions are much more common than others. For instance, since VG has a focus on sports, the triplet \texttt{person holding racket} dominates the bin \texttt{holding}, while \texttt{person holding cellphone} has much lower frequency, although intuitively more common in real world. Instead of dividing triplets based on their predicate type, we propose to divide them based on the frequency of each triplet in training data, which highly correlates with the frequency in test data as well. This way, each bin consists of triplets with roughly the same frequency, and no triplet can dominate others. After computing recall for each bin, we also report the average over bins, which can be seen as a triplet-balanced version of mean recall. Using trial and error, we found the best strategy is to divide bins in logarithmic scale, using powers of 3. This way we will not have too few or not too many bins.

Table~\ref{table:neweval} shows the statistics of each bin in our proposed evaluation setting. Despite the logarithmic scale, we still observe a significant imbalance in the dataset. Specifically, our first bin consists of the rarest triplets, which only appear between 1 and 3 times in training data, and comprise 14.7\% of all triplets in the test set of VG, 75.6\% if we count each unique triplet only once. 
The state of the art~\cite{chen2019knowledge} only achieves
16.7\% recall on that significant portion of data, while achieves 77.1\% recall on the last bin (4 most frequent triplets, 3.1\% of the test set and 0.00004\% of unique triplets). 
This strong disproportion suggests how conventional methods over-invest on few unimportant triplets, at the expense of a large portion of (rare but plausible) real-world situations. 
Accordingly, we believe our new evaluation metric would encourage further research aiming to close the gap.

\section{Qualitative Results \label{sec:qualitative}}

\begin{figure}[t]
\centering
\includegraphics[width=1.0\linewidth]{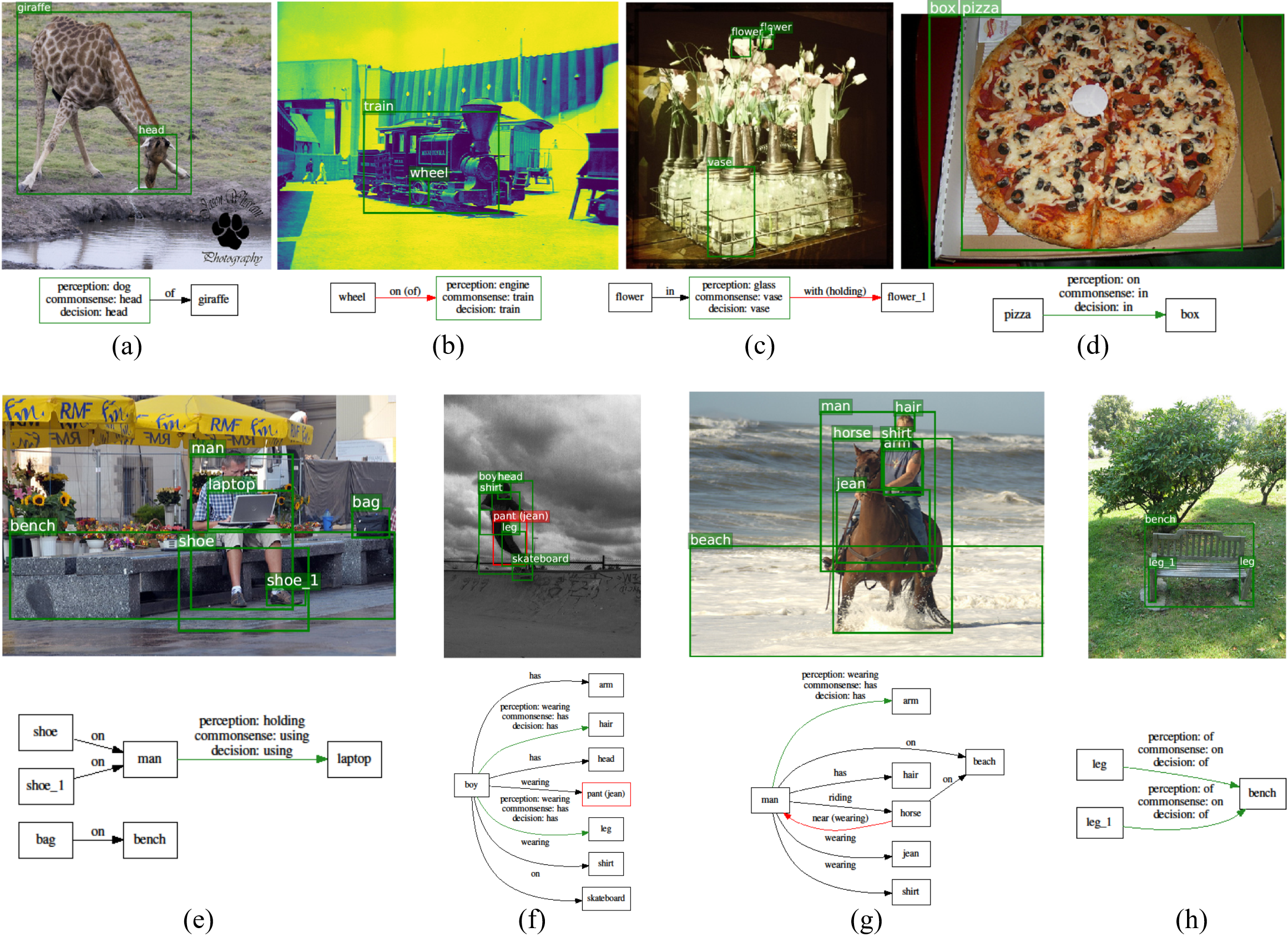}
\caption{Example scene graphs generated by the perception, commonsense, and fusion modules, merged into one graph. Entities are shown as rectangular nodes and predicates are shown as directed edges from subject to object. For entities and predicates that are identically classified by the perception and commonsense model, we simply show the predicted label. But in cases where the perception and commonsense models disagree, we show both of their predictions as well as the final output chosen by the fusion module. We show mistakes in red, with the ground truth in parentheses.}
\label{fig:example}
\end{figure}

To provide more diverse cases of commonsense learned by our model, Figure~\ref{fig:example} shows additional examples of corrections made by our GLAT model to perceived scene graphs. 
In example (a), the perception model mistakes the \texttt{giraffe}'s \texttt{head} for a \texttt{dog}, but our commonsense model corrects that since \texttt{head of giraffe} makes more sense than \texttt{dog of giraffe}. The fusion module correctly prefers the output of the commonsense model, due to its higher confidence. 
In (b), the perception model mistakes the \texttt{train} for an \texttt{engine}, possibly due to the abnormal color palette. Our commonsense model corrects that since \texttt{wheel} more likely belongs to a \texttt{train}.
In (c), the \texttt{vase} has an unusual shape, and is mistaken for a \texttt{glass} by the perception model, also because it is made of \texttt{glass}, but the commonsense model takes into account the fact that the ``\texttt{glass}'' is \texttt{holding} a \texttt{flower}, which is what \texttt{vase}s do.
Moreover, in (d), the \texttt{pizza} is visually \texttt{on} the cardboard, but technically \texttt{in} the box. We, humans, know the latter based on our past experience, and so does the proposed commonsense model. 

In example (e), the \texttt{man} is \texttt{holding} the \texttt{laptop} and \texttt{using} it at the same time. The perception model predicts \texttt{holding} because it is a more visual concept, while the commonsense model predicts \texttt{using} which is more a abstract concept, and in fact a more salient and important verb here. 
In (f), the image is not very clear, and there is no visual distinction between the \texttt{boy}'s body parts and clothing. This makes it hard for the perception model to tell the \texttt{boy} is \texttt{wearing} the \texttt{pants} but \texttt{has} the \texttt{leg}. The commonsense model is robust in such scenarios because it does not rely on the image, but instead considers past experience.
Similarly, (g) makes it hard to distinguish \texttt{man has arm} from \texttt{man wearing shirt}, since the bounding box of \texttt{arm} is highly overlapping with \texttt{shirt}, and there is little visual distinction between their content. Hence, commonsense has a crucial role in distinguishing their interactions by abstracting them into symbolic concepts and ignoring their visual features.
Finally, (h) is a rare case, where the prediction of the commonsense model is wrong, while perception's output was already correct. More specifically, we \textit{miscorrect} \texttt{leg of bench} to \texttt{leg on bench}, because usually things are \texttt{on bench}es in real world. This is while for perception, it is obvious that the \texttt{leg}s are not positioned \texttt{on} the \texttt{bench}. Interestingly, Our fusion module prefers the perceived output this time, and rejects the change made by the commonsense model.

\end{document}